# Deep Learning Models for Visual Inspection on Automotive Assembling Line

Muriel Mazzetto, Marcelo Teixeira, Érick Oliveira Rodrigues, Dalcimar Casanova

Federal University of Tecnology - Paraná (UTFPR), Paraná, Brazil

*Abstract*— *Automotive manufacturing assembly tasks are built upon visual inspections such as scratch identification on machined surfaces, part identification and selection, etc, which guarantee product and process quality. These tasks can be related to more than one type of vehicle that is produced within the same manufacturing line. Visual inspection was essentially human-led but has recently been supplemented by the artificial perception provided by computer vision systems (CVSs). Despite their relevance, the accuracy of CVSs varies accordingly to environmental settings such as lighting, enclosure and quality of image acquisition. These issues entail costly solutions and override part of the benefits introduced by computer vision systems, mainly when it interferes with the operating cycle time of the factory. In this sense, this paper proposes the use of deep learning-based methodologies to assist in visual inspection tasks while leaving very little footprints in the manufacturing environment and exploring it as an end-to-end tool to ease CVSs setup. The proposed approach is illustrated by four proofs of concept in a real automotive assembly line based on models for object detection, semantic segmentation, and anomaly detection.*

## I. INTRODUCTION

For several years, no compromise between product variety and production quantity has been observed. Two basic types of manufacturing systems were key until 1960: one characterized by a high production volume and low product variation, while the second being the opposite, low production numbers and high product variety. Later, the advent of flexible manufacturing systems provided a better harmonization between the high throughput of parts and the ability to produce larger varieties of products. The concept of a flexible manufacturing system filled the gap between the two original types of systems.

In addition to achieving a compromise between product variety and production volume, the flexible manufacturing system has other objectives, which includes [1]:

- Improved Operational Control: Integrating computer systems into the shop floor enables process control, reducing the need for human communication and providing an infrastructure that reacts quickly to manufacturing plan deviations.

- Direct labour reduction: This is usually achieved by automating existing operation. Automations are separated in three levels: fixed automation, flexible automation, and programmable automation.

- Improved short-term responsiveness: Short-term responsiveness may be the result of processing changes, material delivery, or other engineering issues.

- Improved long-term responsiveness: A flexible manufacturing system must be designed to respond to long-term changes when we consider variations in product volumes, the addition of new processes or new products on the production line.

All the goals of flexible manufacturing arise issues that are inherent to certain computer systems. Systems must be advanced enough to quickly adapt to variations or deviations from the original production schedules, identify and distinguish between the different products being processed on the shop floor, perform physical changes of physical configuration and be able to adapt to new models as they are gradually introduced into the line.

Computer systems employed in flexible manufacturing can be divided in three categories [1]: (1) automation systems (e.g. production line robots), which effectively manipulate component parts, assemble and produce; (2) planning systems (e.g. DES, Petry nets), which are responsible for forecasting and adjusting the flow of components and products on the shop floor, adapting to the desired production needs of management and supply problems [2]; and (3) quality control systems (e.g. cameras and sensors), responsible for automatically inspecting possible production failures or defects in the final product.

Several technologies can be employed in the third previously described category [3]. Systems that use computer vision approaches are one of the most common. They use camera-generated images for automatic (does not require human assistance) or semi-automatic (requires some human assistance) inspection of problems such as compliance or fault detection [4].

However, conventional vision systems are strictly dependent on fairly controlled environments, which makes them unsuitable for flexible manufacturing. Being flexible demands constant and rapid reconfiguration of the production line in order to shift from a product to another, or to include new products [5]. In practice, these are severe limitations for CVSs because they have to quickly adapt to texture and luminosity variations, new templates, and dynamic analysis [6]. For a CVS to work under these circumstances, it would require tuning several parameters within a short time. As this is a manual task, inspection can be poor or unfeasible to be done automatically [7]. Also, the efficiency of a CVS depends on robust lighting control, which implies the requirement to halt the production line to fix work pieces for images acquisition. This results in an expensive solution as it increases the operating time over each work piece, the so-called *cycle time*.

The difficulty in adapting computer vision systems to new products increases the cost of end products. Unfortunately, it is necessary to wait until the computer vision system is redesigned in order to properly function on the production line [6]. This automated vision system can often be replaced by a human operator until the new solution is adequate for the task [7].

In recent years, great strides have been made in computer vision systems, especially in systems called end-to-end models, which are based on Deep Learning. Unlike traditional feature-based engineering computer vision systems, this new approach is based on feature learning [6, 8]. Roughly, these models no longer require highly skilled labour to adapt the computer system to new products or tasks in a flexible manufacturing plant. Nowadays, applications of deep learning include translation, speech and audio processing, social network analysis, healthcare and visual data processing [9, 10].

This work proposes a prototype implementation of end-to-end models in a car production line, demonstrating its virtual advantages over traditional models of computer vision, both in the quality of solution and its facilitated adaptability. Furthermore, possible applications for automated inspection (compliance checking and fault detection) are designed based on the proposed vision system.

## II. FEATURE ENGINEERING VERSUS FEATURE LEARNING

Historically, the performance of machine learning methods, including computer vision methods, relies heavily on data representation (also called features or characteristics) [11, 9]. The traditional computational approach is described by 4 basics steps [12]: (1) image acquisition; (2) pre-processing (e.g., correcting for brightness, contrast, colour saturation, etc.); (3) feature extraction (i.e., representing the original data in a reduced dimensional space) and; (4) classification (e.g., contour comparison with a template).

All the steps listed above entail their inherent difficulties, and the solution depends heavily on the addressed problem. No traditional computer vision approach to date has been generic enough to succeed in any given situation. Steps (2) and (3) are especially complicated as features and pre-processing filters must be differently adjusted for each application case, i.e., a set of computational methods is used to identify human faces, but the same set is not adequate to identify flaws in a machined industrial surface. This has always been a criticism of the computer vision research area as each problem requires a unique solution [11].

In this sense, much effort to deploy computer vision algorithms is directed towards the design of the pre-processing and feature extraction pipelines to result in a data structure that is better suited for machine learning [9]. This feature engineering is important, but it requires a lot of work, highly skilled people, and highlights the weakness of current learning algorithms: their inability to automatically extract and organize discriminatory information from data [6].

Deep learning is emerging as a high-tech concept to tackle this problem and it has been adopted by giant organizations [9, 13]. In addition to the high hierarchical structure, deep learning differs from traditional computer vision methods as it is able to suppress, or at least dramatically reduce, pre-processing and feature extraction requirements in traditional methods. That is, these models comprise feature learning capabilities and no longer require any engineering or specialized personnel to perform this task [8].

The feature engineering process of deep learning methods is accomplished by selecting different kernels or adjusting the parameters through end-to-end optimization [9]. Its deep architecture of neural networks with several hidden layers is essentially composed of multilevel nonlinear operations. It transfers the representation or characteristic of each layer to a more abstract upper layer representation. Features such as edge, outline, and object parts are generalized layer over layer. These general representations are written as predictive models that perform classification or regression tasks. In summary, deep learning is an end-to-end learning structure that requires minimal human inference where parameters are adjusted by the algorithm during its training phase.

Traditional computer vision systems perform feature selection separately. Handcrafted features are extracted first where raw data is converted to a different domain (e.g., statistical, frequency, and time domains) in order to obtain representative information that requires specialized domain knowledge. Next, feature selection is performed to improve relevance and reduce redundancy. Traditional neural network techniques are often structured as shallow models containing a maximum of three layers (e.g., input, output, and a hidden layer). Thus, the performance of the model relies not only on the optimization of the adopted algorithms (e.g., multilayer neural network, support vector machine and logistic regression) but is also strongly affected by the handcrafted features. Generally, feature extraction and selection is time consuming and relies heavily on domain knowledge.

In other words, high-level abstract representation in feature learning enables deep learning to be more flexible and capable to adapt to a large variety of data. As data abstraction is considered, the various data types and sources provide no strong influence on the results. On the other hand, the deep hierarchical structure in deep learning enables easier modelling of the nonlinear relationships compared to the superficial structure that is considered in traditional machine learning. In the context of big data in flexible manufacturing, the ability to avoid feature engineering is considered a major advantage due to the challenges associated with this process.

### III. COMPUTER VISION APPLICATIONS FOR AUTOMATED MANUFACTURING INSPECTION

This section covers three different applications of deep learning models for quality inspection in an automotive production line. Methods for object detection, semantic segmentation and anomaly detection are discussed. Furthermore, a proof of concept of the chaining of two methodologies is studied aiming to improve the quality inspection in real time. The experiments were performed in assembly and machining lines of Renault do Brazil with minimal environmental influence as a goal, especially avoiding to change any factory cycle.

3.1 Object Detection

In a classification task, an object is assigned to one of the predefined classes. Several different classification tasks can be found in the domain of quality control such as: classification of an image to determine the presence or absence of a specific component; classification between normal and abnormal configurations; classification of components according to their descriptive features, among others [6, 8, 14].

The majority of the available deep learning classifiers use convolutional neural networks with a varying number of convolutional layers followed by fully connected layers. The availability of manufacture data is limited as compared to the natural image datasets, which drove the development of deep learning techniques in the last 5 years. Therefore, many applications of deep learning in manufacture image classification have resorted to techniques meant to alleviate this issue: the transfer learning. The transfer learning strategy, which involves fine tuning of a network pre-trained on a different dataset, has been applied to a variety of classification tasks.

As mentioned earlier, the main advantage of end-to-end is the fact that no choices are required to be made when it comes to features that should be extracted (as previously mentioned, the classification is performed directly, without the need for any intermediate step simply informing the desired input and output). Figure 1 shows the data used in our case study in a task of classifying an image as disk brake or disk calliper. The images refer to the *ground truth*, i.e. the templates used for supervised learning.

Detection is a task of locating and highlighting (e.g., using a rectangular box) an object in an image. In manufacture, detection is often an important step in the quality control process, which identifies a component or a region of interest for further classification or segmentation [15].

The most common approach to detection for 2-dimensional data is a 2-phase process that requires training of 2 models. The first phase identifies all candidate regions that may contain the object of interest. The requirement for this phase generates high sensitivity and therefore generally produces many false positives [16]. A typical deep learning approach to this is a regression network for bounding box coordinates based on architectures used for classification [17]. The second phase is simply the classification of the subimages extracted in the previous step. The classification step when using deep learning is usually done using transfer learning.

A second approach to detection is a single-phase detector that eliminates the first phase of region proposals. Examples of popular methods that were first developed for detection in natural images and rely on this approach are: You Only Look Once (YOLO) [18], Single Shot MultiBox Detector (SSD) [19] and RetinaNet [20].

The end-to-end model simplifies all the development process as just the desired input and output are required. Figure 1 exemplifies 3 input images with desired delimitation for object detection (ground truth) for the case study presented in this paper. The ground truth consists of highlighting the the position where the piece is located where colors represent different classes.

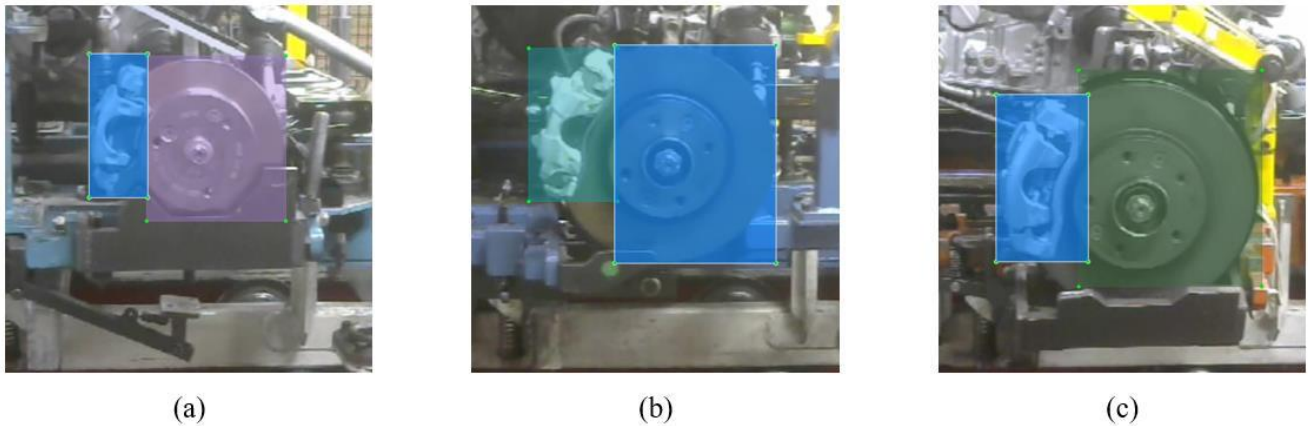

*Fig.1: Ground truth for brake kit object detection. (a) Brake Calliper and Disc 1. (b) Brake Calliper and Disc 2. (c) Brake Calliper and Disc 3.*

Two distinct tasks, brake calliper identification and brake disc identification were analysed using the same architecture. The goal was to use a rectangular box to circumvent the set of pixels that represents each component.

The CVSs proposed in this paper use an *Application Programming Interface* (API) for object detection. The API is integrated to the *Tensorflow* framework [21] and it focuses on measuring and comparing different object detection architectures concerning memory usage, processing speed, and accuracy [22]. The following subsections provides some insights on the used architecture, the case study environment and obtained results.

3.1.1 Object Detection Architecture

This paper integrates *Single Shot Detection* (SSD) [19] and *MobileNet* approaches [23]. MobileNet acts as a convolution network to extract features from image and SSD is responsible for objects scanning [22]. These choices are justified mainly because SSD and MobileNet are suitable for projects that involve hardware limitations such as real-time mobile detection applications [22].

As the proposed architecture is based on CNNs and supervised machine learning, the learning process consists of initially inserting an input image into the CNN, conducting all the convolution and feature extraction operations until the stage of object detection. Next, the error rate is calculated concerning the training set, also called *loss*. The loss value is used to recalculate the CNN weights and the convolution filters, this time following the opposite flow of the architecture.

Every detection is also associated with a *probability* that refers to the percentage of certainty for a model to detect an object. A threshold can be defined for the algorithm to mark a detection only in cases within a specific certainty.

SSD designs bounding boxes of different sizes and shapes throughout the image. In classification, this architecture assigns a percentage of detection according to the presence of objects in the bounding boxes and adjusts their respective sizes and positions to match the layout of the identified objects. At the end of the network, a *Non Maximum Suppression* (NMS) algorithm is applied to eliminate bounding boxes redundancies on the same object. The seminal work [19] uses the CNN *VGG-16* [24] as the basis for feature extraction. However, the API provided by [22] disintegrates SSD and *VGG-16* in order for other architectures to be combined. Figure 2 shows the architecture of the algorithm proposed in [19] that analyses 300×300 images.

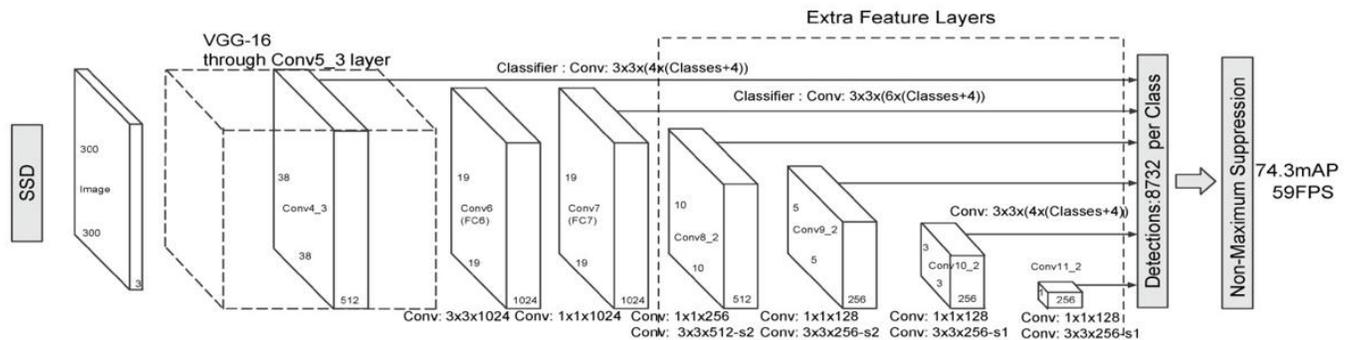

*Fig.2: SSD 300 architecture.*

The foundation for feature extraction in this implementation, the *MobileNet* [23] architecture, replaced *VGG-16* by reducing in 1/30 the computational cost and model size [22]. This architecture is composed by *depthwise separable convolutions*, i.e., it splits a standard convolution into convolution filters for each input channel, and then it applies a 1×1 convolution, called a *pointwise convolution*, to match the output of the *depthwise* convolution.

The efficiency of the proposed object detection model is then assessed. We use the *Intersection over Union* (IoU) rate to express whether or not a detection is correct, depending on the bounding box location concerning the image [22]. The IoU is calculated as in Eq. 1, where *AD* is the box area detected by the trained model; *AG* is the box area marked by the training set; and *IA* is the intersection area between *AD* and *AG*. The closer to 1 the better is the object detection.

$$IoU = \frac{IA}{(AD + AG - IA)} \quad (1)$$

It is also important to define the probability value for accepting detection decisions, as this influences the value of *AD*. However, this metric does not consider cases of incorrect classification or non-classification at all by itself. Therefore, to validate the detection result after its integration to the production line, i.e., assessing the efficiency of trained models, the indexes of *Precision* and *Recall* are used.

*Precision* (Eq. 2) determines how correct the model is in detecting and classifying objects, while *Recall* (Eq. 3) determines the relation between success and failure of the classification. These two metrics are derived from the number of *True Positive* (TP), *False Positive* (FP) and *False Negative* (FN), such that:

- TP is every correct classification for which the bounding box has a IoU more than 0.5;
- FP combines detections with IoU less or equal to 0.5; detection of classes that are not part of the original image; or overlapping detection, which has already been counted as TP;
- FN is all unrealized classification, i.e., images where the model was unable to create the bounding box correctly.

$$Precision = \frac{TP}{(TP + FP)} \quad (2)$$

$$Recall = \frac{TP}{(TP + FN)} \quad (3)$$

Using the concepts presented so far, the next subsection discusses the application of the multiple object detection model in the automotive industry.

### 3.1.2 Object Detection Case Study

The following experiment was conducted on a real automotive assembly line, at the Renault do Brazil vehicle factory. The selected case concerns to the brake disc and calliper set conformity and anomaly detection.

In the assembly line used in this experiment, *Automatic Guided Vehicles* (AGVs) are loaded with workpieces for all types of products to be manufactured throughout the line. As humans conduct this task, it is error-prone, and kits can be assembled with components belonging to different vehicle models. To mitigate this problem, we propose a detection system that verifies whether the kit is assembled with appropriate workpieces. Tests were performed with three types of brake discs and three types of brake callipers, summing up to six classes. Figure 3 shows examples of sets and their respective classes.

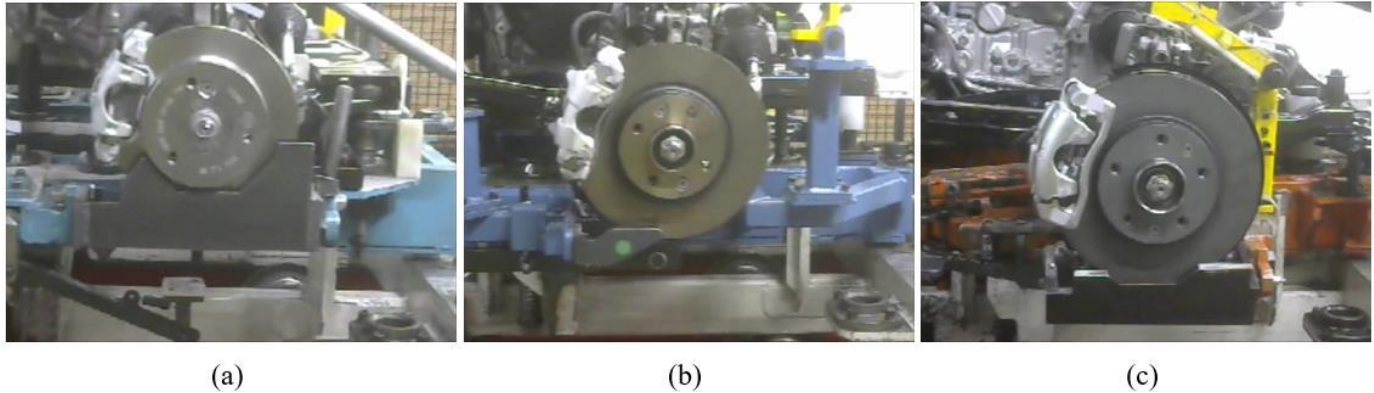

*Fig.3: Types of brake disc and calliper mounted on track support. (a) Brake calliper and Disc 1. (b) Brake calliper and Disc 2. (c) Brake calliper and Disc 3.*

Objects demonstrate very low variation due to the standardization of factory environments. In other words, virtually no variation in shape or colour is observed between pieces of the same type. Due to the apparent homogeneity, few images were used for training the classification model: a total of 20 images (400×400 pixels) of each class. Testing sets were composed of 15 images (400×400 pixels) for each class. This configuration produced a classification model that is evaluated in the following section. Figure 4 shows the learning convergence of this model.

### 3.1.3 Object Detection Results

A total of 321 new images were collected from video frames captured on the assembly line. Table 1 presents the results for six evaluated classes, with probability threshold of 90%. The threshold values used for the detection probability were 60%, 65%, 70%, 75%, 80%, 85%, 90% and 95%. Figure 5 shows *Precision* and *Recall* of analysed classes over various probability thresholds.

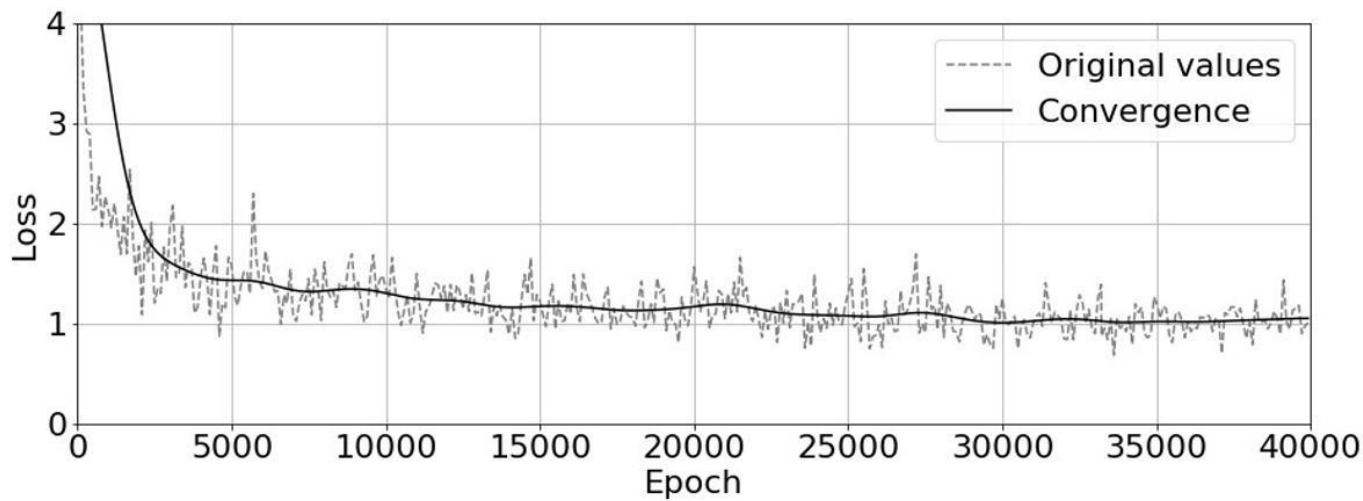

*Fig.4: Learning convergence (i.e., loss as a function of steps) for brake disc and calliper detection model.*

Table 1: Disc and calliper detection in images

|  | Disc | | | Calliper | | |
|---|---|---|---|---|---|---|
| Type | 1 | 2 | 3 | 1 | 2 | 3 |
| TP | 79 | 66 | 50 | 89 | 66 | 49 |
| FP | 0 | 0 | 0 | 0 | 0 | 0 |
| FN | 35 | 26 | 23 | 8 | 11 | 12 |
| *Precision* | 1.00 | 1.00 | 1.00 | 1.00 | 1.00 | 1.00 |
| *Recall* | 0.69 | 0.72 | 0.68 | 0.91 | 0.86 | 0.80 |

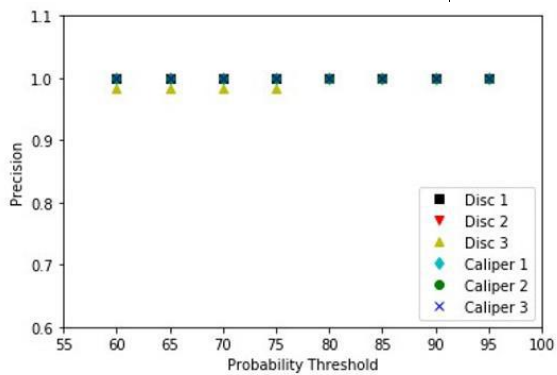
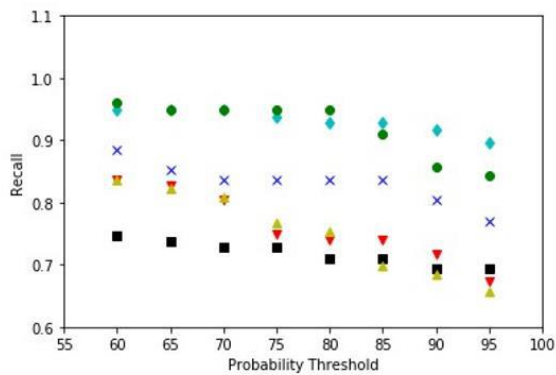

(a)      (b)

*Fig.5: (a) Precision × Probability threshold. (b) Recall × Probability threshold.*

It is important to note that in Figure 5a, the higher the detection probability threshold, the higher is the *TP* rate of the model for all analysed classes, due to the low standard deviation. In contrast, in Figure 5b, the higher is the amount of *FN*, i.e., of unclassified work pieces.

When *FN* cases were analysed, we found out that they were related to images occluding part of the unclassified component due to improper video capture, as illustrated in Figure 6.

In summary, the approach used to detect images using video frames introduces several benefits to the assembly line. The possibility of detecting objects without halting the production line is one of these benefits. A side effect of this flexible approach is capturing frames at the beginning or the of the work piece path, as shown in Figure 6.

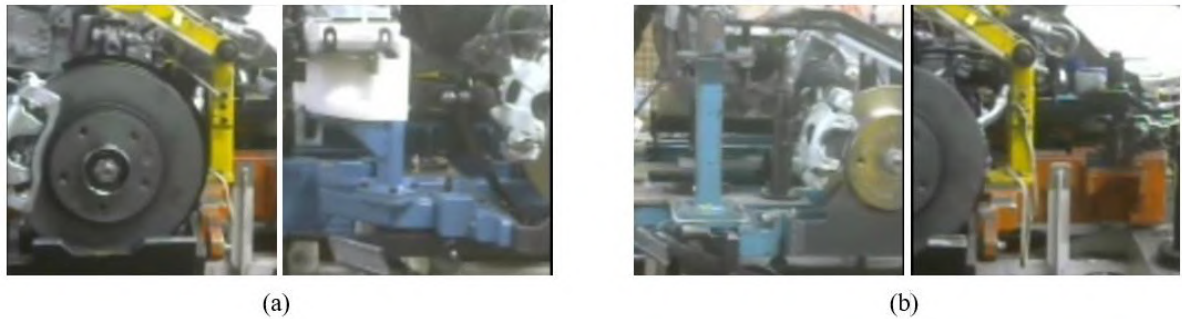

*Fig.6: Samples of unclassified images, which generated FN. (a) Model was unable to classify the calliper. (b) Model was unable to classify the disc.*

However, this effect can be easily overcome when temporal configuration is applied. In this work, a trigger was configured to select classes during phases where work pieces pass in front of the camera. Thus, the final acquisition is based on the classes that were identified in more frames during that period, chosen as vote by the majority.

New 720×720pixels, 30 fps videos were captured to test this approach. The analysed video segment refers to the phase when the work pieces pass in front of the camera, where the first successful detection is the initial trigger. The process is stopped 100 frames without any detection.

Figure 7 displays excerpts from one of the test. Figure 7a displays the frame where the first detection is initialized. Figure 7b shows the moment when the kit is centralized and Figure 7c displays a frame that produced a FN in brake disc detection. At last, Figure 7d shows the termination of the tracking after 100 consecutive frames without detection.

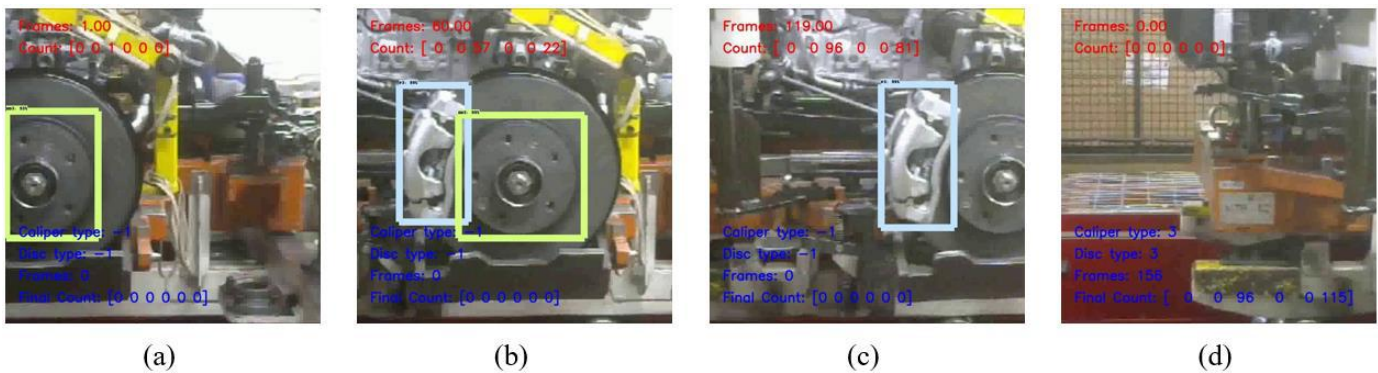

*Fig.7: Counting detection of brake disc and calliper mounted on track support. (a) Start of count. (b) Half of the detection path. (c) Frame with FN in brake disc detection. (d) End of count.*

Three videos were used to evaluate each set of assembled disc and calliper, which sums up to nine videos that were used to test the entire proposal. Table 2 shows the amount of classes for each video considering a 95% probability threshold for detection. In this table, videos 1-3 were assembled with type 1 disc and calliper; videos 4-6 represent type-2; and videos 7-9, type 3. The amount of frames per video varied along the progress of the production line.

Table 2 shows a 100% of accuracy in the detection task. Therefore, this approach enables mitigation of the *FN* problem that leads to incorrect detections, as it does not bind the final decision to a single image, but to the vote by the majority within the analysed video segment.

Although it is an accurate tool for locating objects and classifying them, it is not yet able to classify the boundaries of each analysed object, or to categorize different regions of the object. A proof of concept using semantic segmentation was performed to address this idea, which is described in the next section.

*Table 2: Disc and calliper detection in videos*

| Video | Disc type: 1 | 2 | 3 | Caliper type 1 | 2 | 3 |
|---|---|---|---|---|---|---|
| 1 | 113 | 0 | 0 | 119 | 0 | 0 |
| 2 | 105 | 0 | 0 | 118 | 0 | 0 |
| 3 | 103 | 0 | 0 | 103 | 0 | 0 |
| 4 | 0 | 118 | 0 | 0 | 115 | 0 |
| 5 | 0 | 117 | 0 | 0 | 116 | 0 |
| 6 | 0 | 115 | 0 | 0 | 112 | 0 |
| 7 | 0 | 0 | 96 | 0 | 0 | 115 |
| 8 | 0 | 0 | 112 | 0 | 0 | 120 |
| 9 | 0 | 0 | 108 | 0 | 0 | 119 |

### 3.2 Semantic Segmentation

In image segmentation tasks, an image is divided in different regions to separate distinct parts or objects [25]. In manufacture, the common applications are segmentation of parts that are often addressed as a pre-processing step for feature extraction and classification [26].

The most straightforward and still widely used method for image segmentation is classification of individual pixels based on small image patches (both 2-dimensional and 3-dimensional) extracted around the classified pixel. It enables the use network architectures and solutions that are widely known to work well for classification tasks. However, some shortcomings for this approach are [25]: (1) computationally inefficiency, as it processes overlapping parts of images multiple times; (2) each pixel is segmented based on a limited-size context window and ignores the wider context. In some cases, (3) a piece of global information, e.g. pixel location or relative position in relation to other image parts, is required to correctly assign its label.

One approach that addresses the shortcomings of the pixel-based segmentation is a fully convolutional neural network (fCNN) [27, 28, 29]. Networks of this type process the entire image (or large portions of it) at the same time and output a 2-dimensional map of labels (i.e., a segmentation map) instead of a label for a single pixel. Architectures that were successfully used in both natural images and radiology applications are encoder-decoder architectures [30, 31].

The end-to-end model simplifies all the segmentation process. Figure 11 illustrates an input image and the desired output image in a task were a cylinder head image needs to be segmented to identify machined regions. Inspection of machined regions is especially important as it must be free of defects. Otherwise, these defects can result in loss of power, leakage and engine malfunction.

This paper uses a tensorflow API for semantic segmentation [32], called Deeplab V3+, that combines the benefits of spatial pyramid pooling module and encode-decode structure to increase segmentation task results [26]. The following subsections detail the architecture used, the case study environment and the results.

#### 3.2.1 Semantic Segmentation Architecture

The architecture for semantic segmentation used in this paper is the Deeplab V3+, which applies several parallel atrous convolution with different rates, called Spatial Pyramid Pooling (ASPP), to capture the contextual information at multiple scales [26].

The Deeplab V3+ architecture employs the spatial pyramid pooling module (Figure 8a), with the encoder-decoder structure (Figure 8b). DeepLabv3+ contains rich semantic information from the encoder module, while the detailed object boundaries are recovered by the simple yet effective decoder module. The encoder module allows the extraction of features at an arbitrary resolution by applying atrous convolution (Figure 8c).

The DeepLabv3+ employs an encoder-decoder structure. The encoder module encodes multi-scale contextual information by applying atrous convolution at multiple scales, while the simple yet effective decoder module refines the segmentation results along object boundaries. Figure 9 shows the details of the encoder-decoder structure.

Differing from the conventional 3×3 depthwise separable convolution (Figure 10 a), it applies a single filter for each input channel alongside a pointwise convolution (Figure 10b), which combines the outputs from depthwise convolution across channels, the atrous separable convolution with rate = 2 (Figure 10c) takes pixel information by adding an interval of one pixel in the stride size. When the rate is altered, it is possible to explicitly control the resolution of features computed by deep convolutional neural networks and adjust the filter field-of-view to capture multi-scale information [26]. It is important to note that a standard convolution is the same as an atrous convolution with rate = 1.

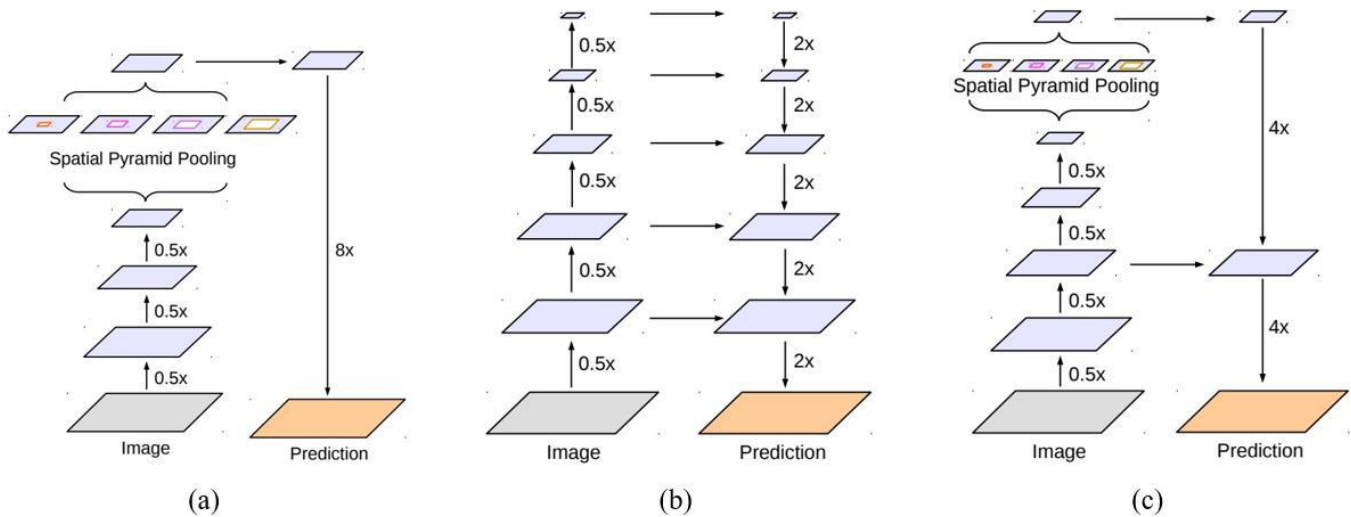

*Fig.8: Deeplab V3+ components representation [26]. (a) Spatial Pyramid Pooling; (b) Encoder-Decoder; (c) Encoder-Decoder with Atrous Conv.*

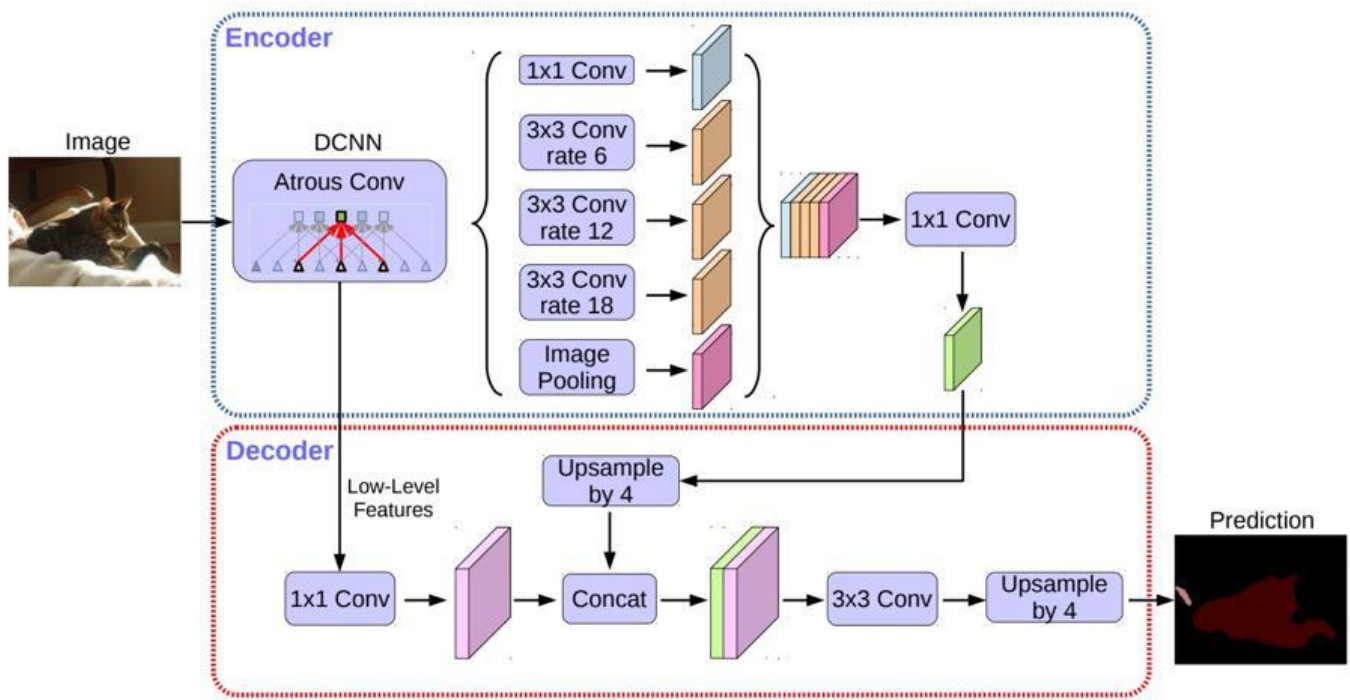

*Fig.9: Encoder-Decoder structure [26].*

As supervised learning, each image must have an equivalent label map (also called *ground truth*). The creation of this map consists in colouring specific regions of the image with pre-established colours that represent the classes in images.

The model evaluation is performed by extending the IoU (Eq. 1), presented in section 3.1, to all image pixels. The API already provides an evaluating tool based on the IoU [32].

Next section uses concepts covered here, applied in a proof of concept to segment regions of the image in a machining line, also called semantic segmentation.

3.2.2 Semantic Segmentation Case Study

The case study approached with semantic segmentation, using the Deeplab V3+ API [32], was carried out in a real cylinder head machining line at the Renault do Brazil engine factory. This methodology was addressed to generate masks to evaluate part quality at the end of the production line, leveraging the operator focus to other tasks. Model training was performed using a *GeForce GT 540m GPU* with 2 Gigabytes of memory.

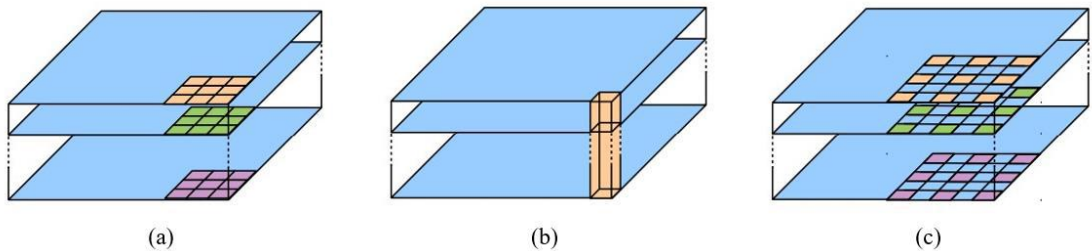

*Fig.10: Example of kernel's dimensions for convolution [26]. (a) Depthwise convolution; (b) Pointwise convolution; (c) Atrous convolution.*

This performed test is just a proof of concept (PoC), with no intent to focus on maximizing hit rates or optimizing training. The general objective of this study is to verify the applicability of deep learning semantic segmentation as an assistant to the inspection task.

The proposed system consists of capturing the image of the face of a piece and generating masks in different regions.

Moreover, in the proposed proof of concept, the model was trained with samples of defective parts to generate suggestions of regions with issues. Figure 11 shows an example of a cylinder head on the inspection table next to its respective label map.

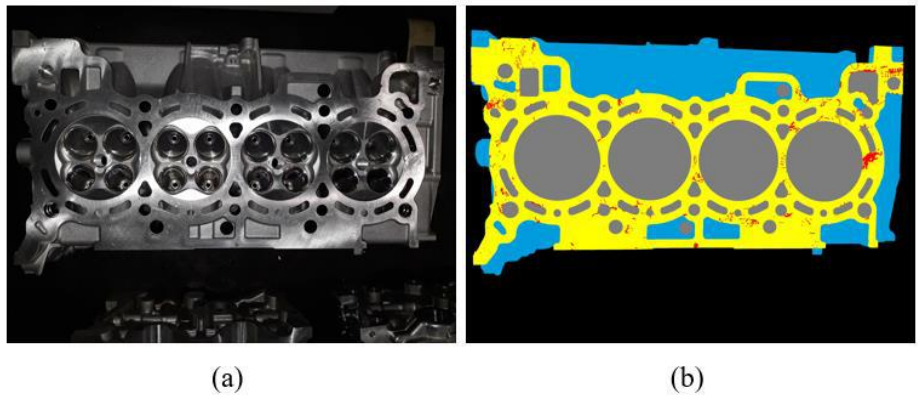

*Fig.11: Example of cylinder head images used in training dataset. (a) Cylinder head on the evaluation table. (b) Cylinder head label map.*

It is possible to observe in Figure 11b the following labels: (1) black is used to label everything that does not belong to the cylinder head, i.e., background; (2) blue symbolizes regions of the gross cylinder head, i.e., without machining process; (3) yellow represents the machined surface of the cylinder head; (4) grey is used to label holes; and finally, (5) red was used to label defects in the part.

It is important to note that these defects were synthetically generated in scrap parts, and hence they are not part of the actual defects of the production line. These defects were synthetically generated for the presented proof of concept.

A total of 10 manually catalogued images were used. The images were captured by a smartphone (4128×3096 pixels). Eight images were selected to compose the training dataset and 2 to compose the test dataset. Each image was divided in 36 patches to force losing as little information as possible when resizing the input dimensions, which results in 688 × 516 images. After this patch separation, 288 images were used for training and 72 for testing. The model was trained for 1000 steps with a size 2 mini-batch due to restrictions of the video memory.

When it comes to architecture parameters, the atrous rates were 6, 12 and 18. The output stride was set to 16, and the decoder output stride to 4. The input images were resized to 321 × 321 pixels in order to match the input from Deeplab V3+ architecture. The base architecture selected to feature extraction is Xception 65.

The algorithm flow consists of receiving an input image, dividing it into 36 image patches, evaluating each patch and reconstructing the final image with the output of each semantic segmentation. The results of this experiment are described as follows.

### 3.2.3 Semantic Segmentation Results

As mentioned earlier, the cylinder head images were evaluated at the end of the production line. Images were captured after removing the parts from the conveyor, over a table where they are checked by an operator to catalogue them as good or bad parts.

The face selected for analysis is considered the most important because it is the contact face with the engine block gasket and cannot contain scratches, porosity or excess material as they may cause leakage or obstruction of cooling and lubrication. Besides, it is also the face that has the most contact with the conveyor during fabrication, which may result in scratches of chip rubbing.

The proposed CVS PoC aims to assist the operator by segmenting the regions to be evaluated according to an evaluation sequence, masking the other regions in each step. Besides, the system also aims to present possible regions with defects as a suggestion for the operator.

A total of 72 images respective to two cylinder heads were evaluated. Figure 12 displays some of these images along with their respective label maps and segmentation predictions.

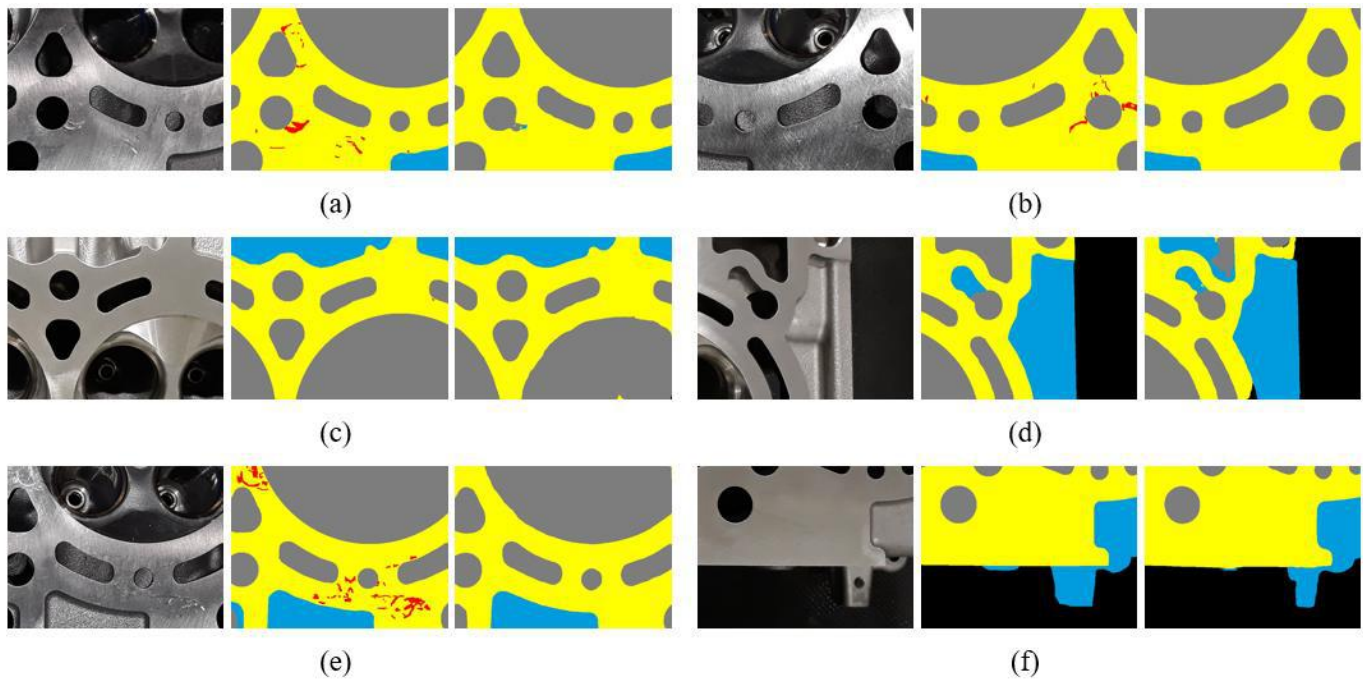

*Fig.12: Original input, label map and semantic segmentation prediction from respective images.*

Notably, as shown Figure 12, the trained model was not able to detect defects. This problem can be related to the low amount of images used for training, affecting the generalization performance (overfitting), and also due to information loss during image resizements. In a continuation of PoC, input images must be treated differently, allowing cropping to be made directly at the architectural input dimensions to take advantage of all pixels of the original image. However, it is still possible to verify that the system is able to create masks on the parts in an acceptable way, strictly following the delimitation of the specified regions.

In addition to visual evaluation, the API provides a *mean Intersection over Union* (mIoU) result based on all test dataset images, taking into account the pixels of the semantic segmentation output relative to the label map. For the proof of concept presented in this article, the result of mIoU was 0.7894.

Although supervised learning requires manual annotation of label maps, which is a laborious function, the proposed system does not require an expert or engineer to inspect the work, the annotations can be performed by non-experts.

### 3.3 Anomaly Detection

A common need for real-world datasets is determining which data point stand out as being different to all others data points. Such data are known as anomalies, and the goal of anomaly detection (also referred to as outliers, novelties, noise, exceptions and deviations) is to identify these anomalous points. Errors in data can produce anomalies but can also be indicative of a new, previously unknown, underlying process.

In other words, anomaly detection is a technique used to identify unusual patterns that do not conform to expected behaviour. It can be considered the thoughtful process of determining what is normal and what is not. Anomaly detection is applicable in a variety of domains such as intrusion detection, health monitoring systems, fraud detection in credit card transactions, fault detection in operating environments, detection of fake news and misinformation over the Internet, industry quality control inspection, security and surveillance.

The major difficulty in this type of computer vision task is that it is nearly impossible to generate a balanced database to train a supervised learning algorithm [14] as anomalous or faulty events barely occur. Due to this fact, semi-supervised and unsupervised approaches are dominant recently. In contrast, semi-supervised and unsupervised methods do not require data labelling (or requires little), being more suited towards rare/unseen anomalous cases.

Generative Adversarial Network (GAN) architecture are the most prominent method in this class of problems, especially in computer vision. Unlike traditional classification methods, the GAN-trained discriminator learns to detect false from real in an unsupervised fashion, which leads GAN to an attractive unsupervised machine learning technique for anomaly detection [33]. Furthermore, the GAN framework produces a generator which is an explicit model of the target system with its ability to output normal samples from a certain latent space.

The GAN architecture was used in two case studies for anomaly detection. One study focused only on detecting cylinder head defects in the engine machining line, and the second focus in evaluating the object detection chaining with brake disk anomaly detection. The architecture, case studies and their results are described in the following subsections.

#### 3.3.1 Anomaly Detection Architecture

This paper uses a state of art deep learning model called *AnoGAN* for anomaly detection, chosen due to the use in recent deep learning methods [34] and associated good results in benchmarks presented by [14].

The architecture used in this article is based on GANs, a framework proposed by [34] that consists of two adversarial modules, a generator *G* and a discriminator *D*. The generator *G* learns a distribution $p_g$ over data *x* via a mapping *G(z)* of samples *z*, 1D vectors of uniformly distributed input noise sampled from latent space *Z*, to 2D images in the image space manifold *X*, which is populated by healthy examples. In this setting, the network architecture of the generator *G* is equivalent to a convolutional decoder that utilizes a stack of strided convolutions. The discriminator *D* is a standard CNN that maps a 2D image to a single scalar value $D(\cdot)$. The discriminator output $D(\cdot)$ can be interpreted as probability that the given input to the discriminator *D* was a real image *x* sampled from training data *X* or generated *G(z)* by the generator *G* [14]. In other words, this method trains a generative model and a discriminator to distinguish real and generated data simultaneously, illustrated by Figure 13.

Let us suppose a trained generator *G* and discriminator *D*, given a new query image *x*, the function $\mu(x) = x \rightarrow z$ proposed by [14] finds a point *z* in the latent space that corresponds to an image *G(z)* that is visually most similar to query image *x* and that is located on the manifold *X*. That is, it seeks to recreate an image considered normal from the characteristics found in the query image *x*.

The anomaly detection score *A(x)* (Eq. 4) consists of a *residual loss* $L_R(z)$ (Eq. 5), that measures the visual dissimilarity between query image *x* and generated image *G(z)* in the image space, and a *discrimination loss* $L_D(z)$ (Eq. 6) calculated by imputing the generated image into the discriminator model, where *σ* is the sigmoid cross entropy with logits *D(G(z))* and targets *α* = 1.

$$A(x) = (1 - \lambda) \cdot L_R(\mu(x)) + \lambda \cdot L_D(\mu(x)). \quad (4)$$

$$L_R(z) = |x - G(z)|. \quad (5)$$

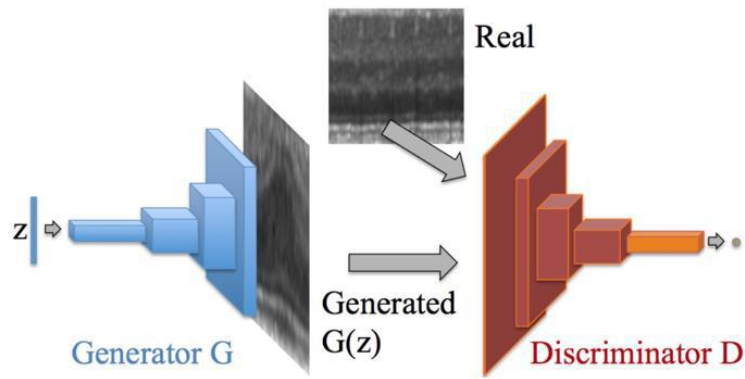

*Fig.13: Deep convolutional generative adversarial network [14].*

$$L_D(z) = \sigma(D(G(z)),\alpha). \quad (6)$$

Additionally, the residual image described by Eq. 7 is used to generate the color map for identification of possible anomalous regions within an image. The colormap selected to this work is the colormap jet, available on opencv.

$$x_R = |x - G(z)|. \quad (7)$$

The architecture used for anomaly detection is based on the Adam optimizer, with a learning rate of 0.00001 and $\beta = 0.1$ for the discriminator model, a learning rate of 0.0002 and $\beta = 0.5$ for generator model; choices that were based in [14]. The latent space vector dimension was set to 100.

The input is set to receive 200 × 200 RGB images. The sequence of anomaly detection process is illustrated in Figure 14, where *model_1* refers to the discriminator model and *model_2* refers to the generator model. Figure 15 and Figure 16 show the architectures of discriminator and generator models, respectively.

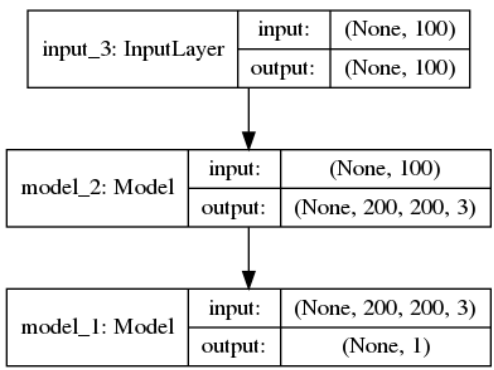

*Fig.14: Anomaly detection process.*

The same training concepts covered in section 3.1 are valid for AnoGAN training aspects, with the exception that it is an unsupervised learning algorithm.

### 3.3.2 Cylinder Head Case Study

The first case study was carried out on a cylinder head machining line, evaluating one of the part faces at the end of the production line. The face selected for analysis is the same mentioned in the semantic segmentation case study.

Although the factory environment has constant artificial lighting, the incidence of external light through the windows varies according to the period of the day, season, etc. Therefore, solutions proposed with conventional CVSs require envonriment enclosure, depriving external light, and dedicated illumination to highlight imperfections, which requires costly adaptations in the conventional plant environment. Also, as it is a specifically designed lighting system, slight variations in part angle result in a lot of reflection.

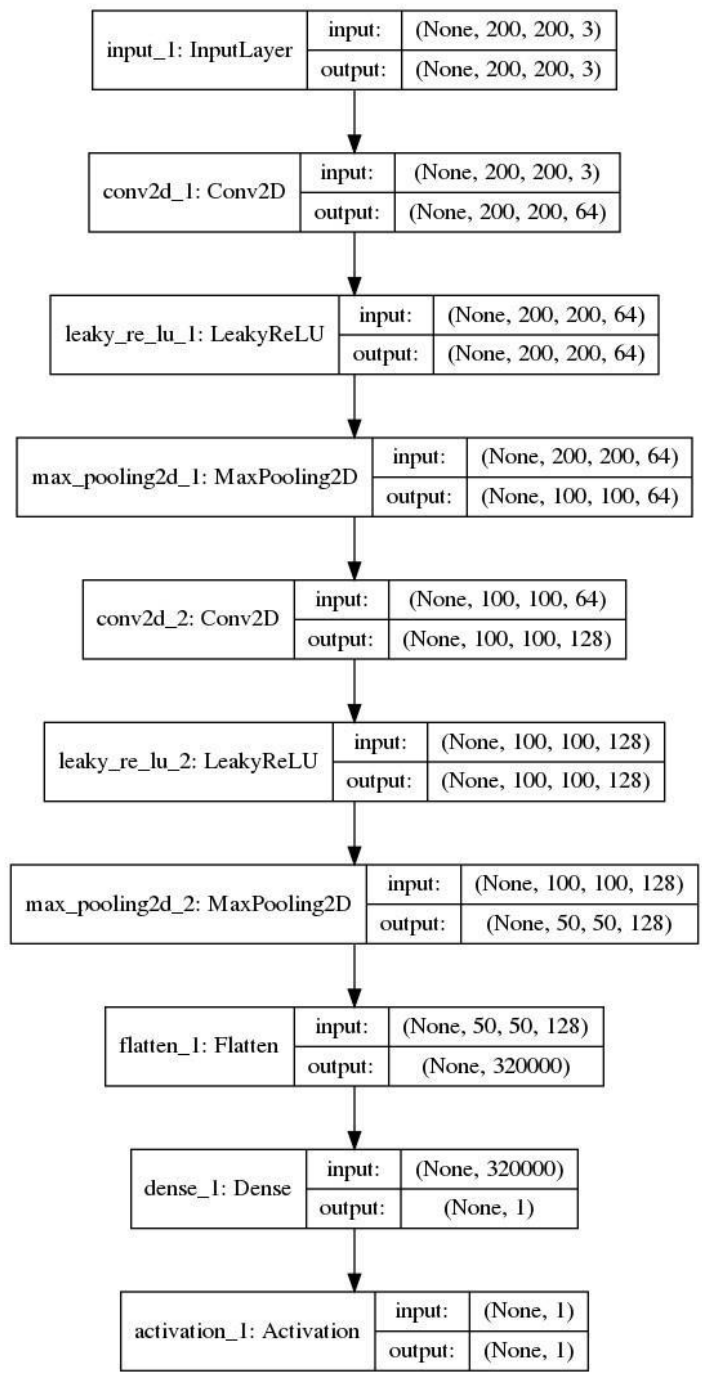

*Fig.15: Discriminator model D architecture.*

To address this problem while being very little intrusive when it comes to the production line and aiming to mitigate the difficulty of analysing parts under different ambient lighting conditions, this paper proposes the use of deep learning anomaly detection.

The proposed system was introduced at the end of the production line, acting during the course in which a robot handles the part. The robot was synchronized with a Basler acA3800 10gc camera, performing centralized image capture on each valve guide. Although it was necessary to change the path of the original robot, the additional triggering operation did not increase the original duty cycle time. Figure 17 shows an example of images captured from a cylinder head.

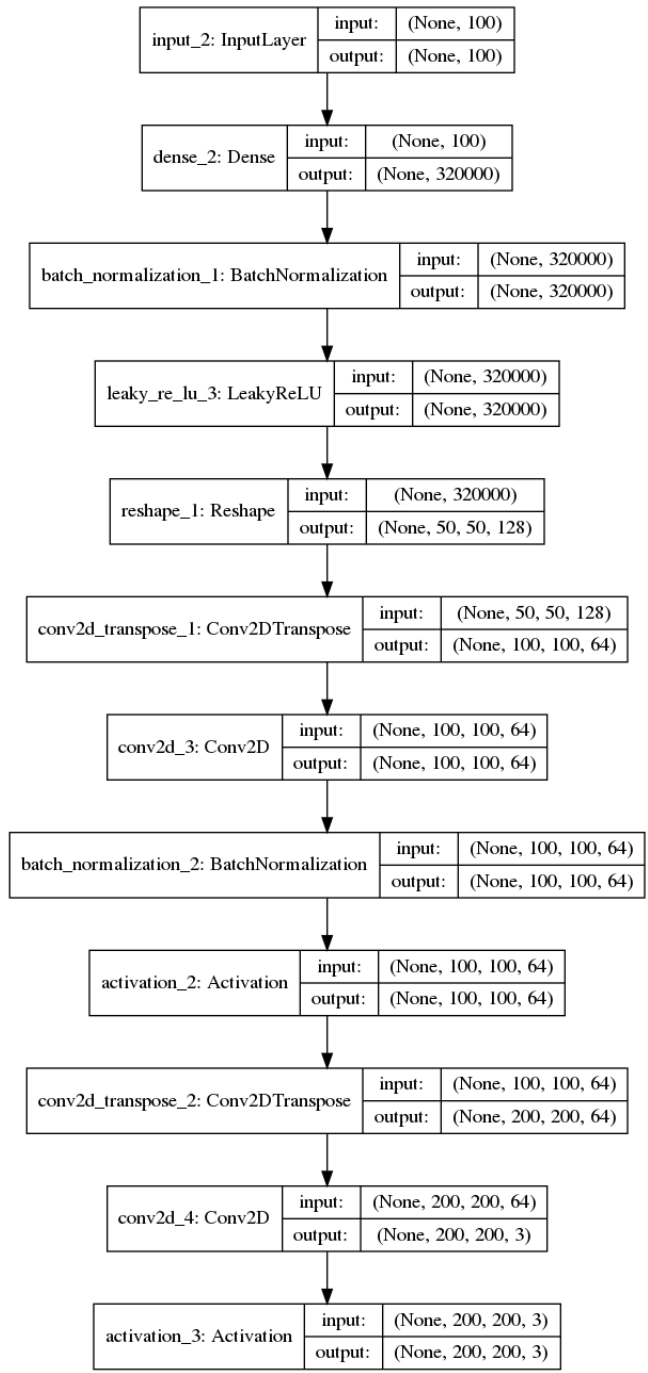

*Fig.16: Generator model G architecture.*

The camera was set to capture 2748×2748 coloured images. A total of 16000 images were selected to train the system, referring to 4000 pieces considered normal according to the quality criteria. All images were resized to 200 × 200 pixels, according to the dimensions specified in the architecture, due to the memory limitation of the used GPU. The system was trained with 10000 steps with a mini-batch of size 16. The following are the results of this case study.

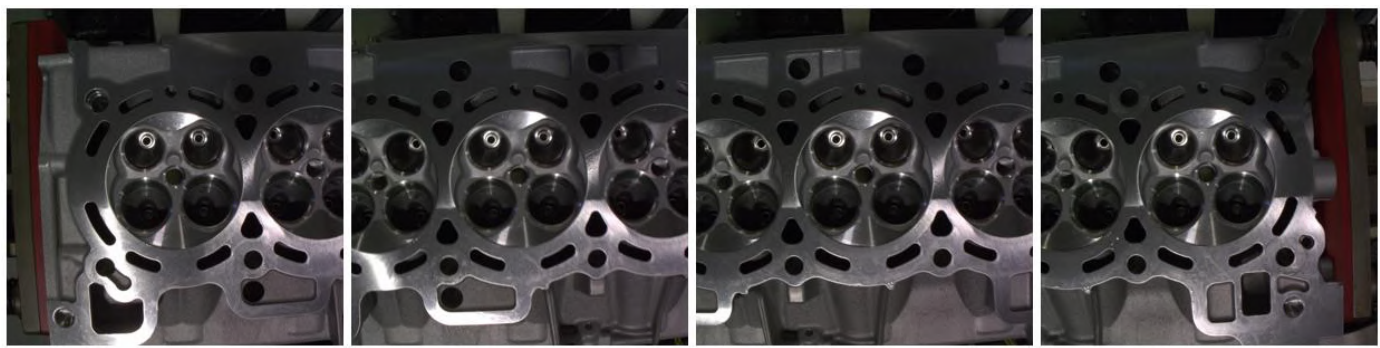

*Fig.17: Examples of acquired pictures of a cylinder head.*

### 3.3.3 Cylinder Head Results

In order to determine if an image contains an anomaly, a value based on Eq. 4 was empirically established by analysing 1000 images considered normal, using the highest anomaly score value of the analysed images as threshold. If the input image produces a score greater than the threshold, a residual image based on Eq. 7 is generated to assist in highlighting the possible anomaly region.

The test of the model in cases with anomaly occurred by evaluating real cases, such as the cases shown in Figure 18.

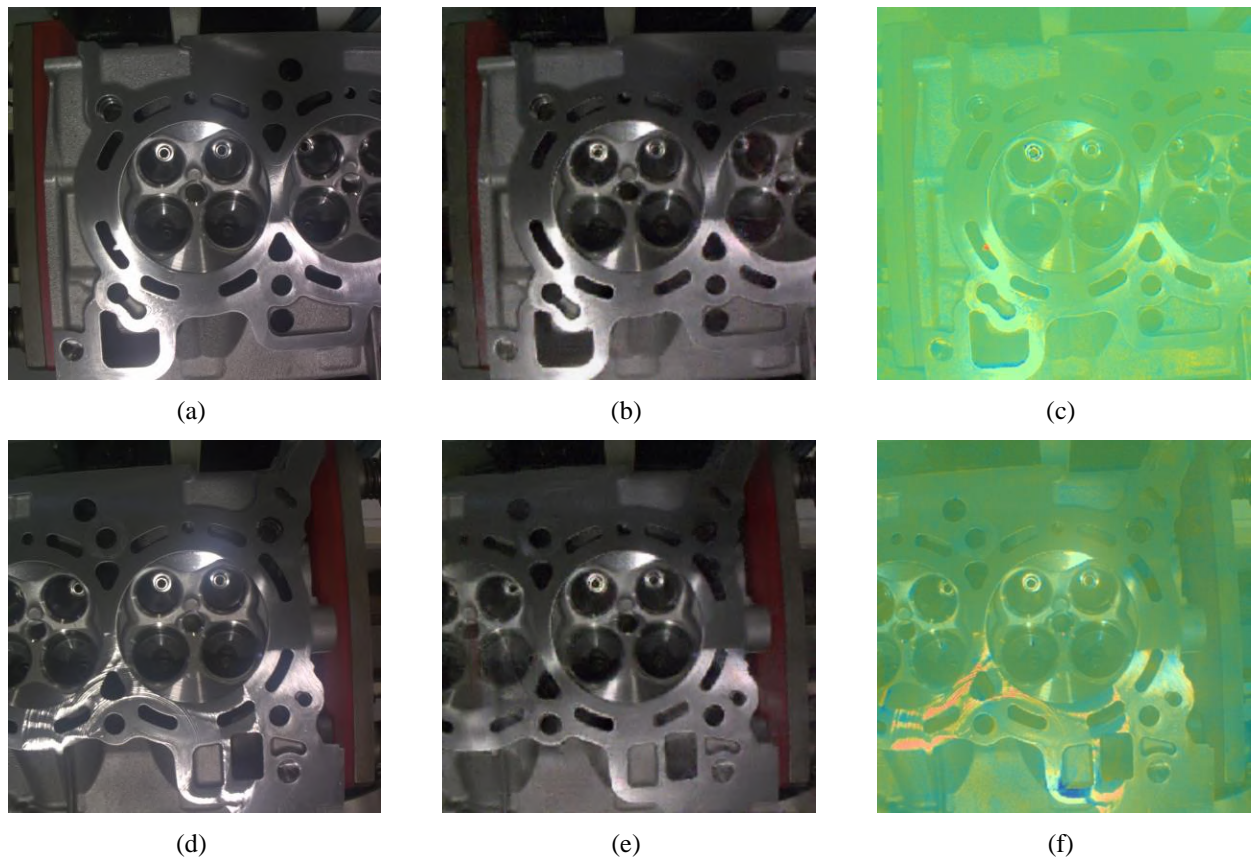

*Fig.18: Anomaly detection in defective parts. (a) and (d) are the real captured images. (b) and (e) are generated image from respective input. (c) and (f) are the residual image with colormap applied over original input.*

The colormap selected to this work is the colormap jet, available on opencv. The colours are represented within a range of -1 to 1, derived from the subtraction between the generated image and the original image. In this representation, the colors closer to green represent normality, whereas red represents anomalies.

It is important to note that what defines the anomaly with greater confidence is the value of the anomaly score. The colormap just to assists with the visualization of potential anomalous regions, as described by [14].

The system is physically unable to differentiate scratches and pores smaller than 1mm due to the scale factor of object dimensions. It was possible to notice the limitation of the model when evaluating pores and scratches in surfaces of the pieces smaller than 5mm$^2$.

Even with these limitations, in a side-by-side test with production line supervisors evaluating the same images, the system was able to indicate all errors pointed out by supervisors. The main advantage of this system over the human operator is the evaluation time, requiring only 3 seconds per image. In addition, the system is low cost and simple to set, while not requiring parameter adjustments.

#### 3.3.4 Brake Kit Case Study

The second case study is addressed to perform anomaly detection on the complete brake kit assembly after checking the conformity of parts. For this purpose, the location of the parts in object detection was used to cut the assembled kit region using the length between the left edge of the calliper location and the right edge of the brake disc as the width and height of cropped image. Only frames containing both object detected were used. All new images were resized to 200 × 200 pixels for standard inputs to the architecture specified. Some examples are shown in Figure 19, related to a type 3 kit detection.

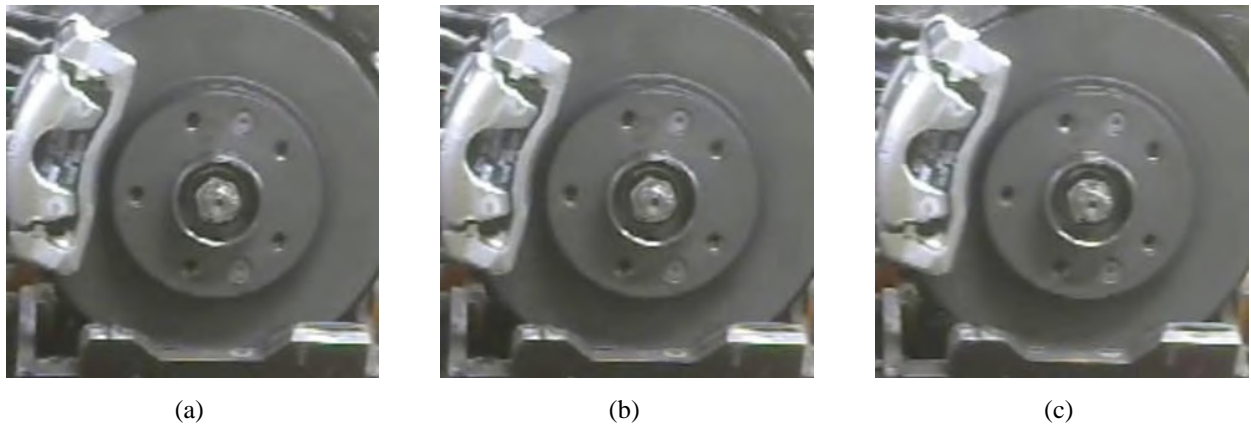

(a)      (b)      (c)

*Fig.19: Cropped frames of brake kit detection: (a) is the first detection in camera's field of view; (b) is in half of the detection path; (c) is in the last detection in camera's field of view.*

A specialist model was trained for each type of brake kit, using 900 images of parts considered normal for each. The corresponding model is triggered by the result of the previous object detection, using the frame with the highest probability detection for anomaly analysis. The parameters used for training were 1.000 steps and mini-batch of size 16. The following subsection describes the results for this case.

#### 3.3.5 Brake Kit Results

In order to determine if the analysed image contains an anomaly, a value based on Eq. 4 was established after analysing 500 new images considered normal from each kit, using the highest anomaly score value of the analysed images as threshold. If the input image produces a score greater than the threshold, a residual image based on Eq. 7 is generated to assist in highlighting the potential anomalous region.

Some instances of normal images and their analyses are shown in Figure 20, where each row concerns to a different type of brake kit. These images show the original input image, its respective image recreated by the generator model and the residual image.

The test of the model in cases with anomaly occurred by evaluating real cases captured on video, such as the case shown in Figure 21. In this occasion, the type 3 disc features a brake calliper type outside of the object trained models, inserted in the manufacturing line to a new car model. Although rejected by the conformity test in the previous step, this case was separated to evaluate the anomaly detection. In Figure 21b it is possible to observe the image created by the generator model, trying to approximate the similar components found in the image of the components considered normal. Once again, it is important to note that what defines the anomaly with greater confidence is the value of the anomaly score, and that the colormap function is to assist with the visualization of potential regions with anomaly, as described by [14].

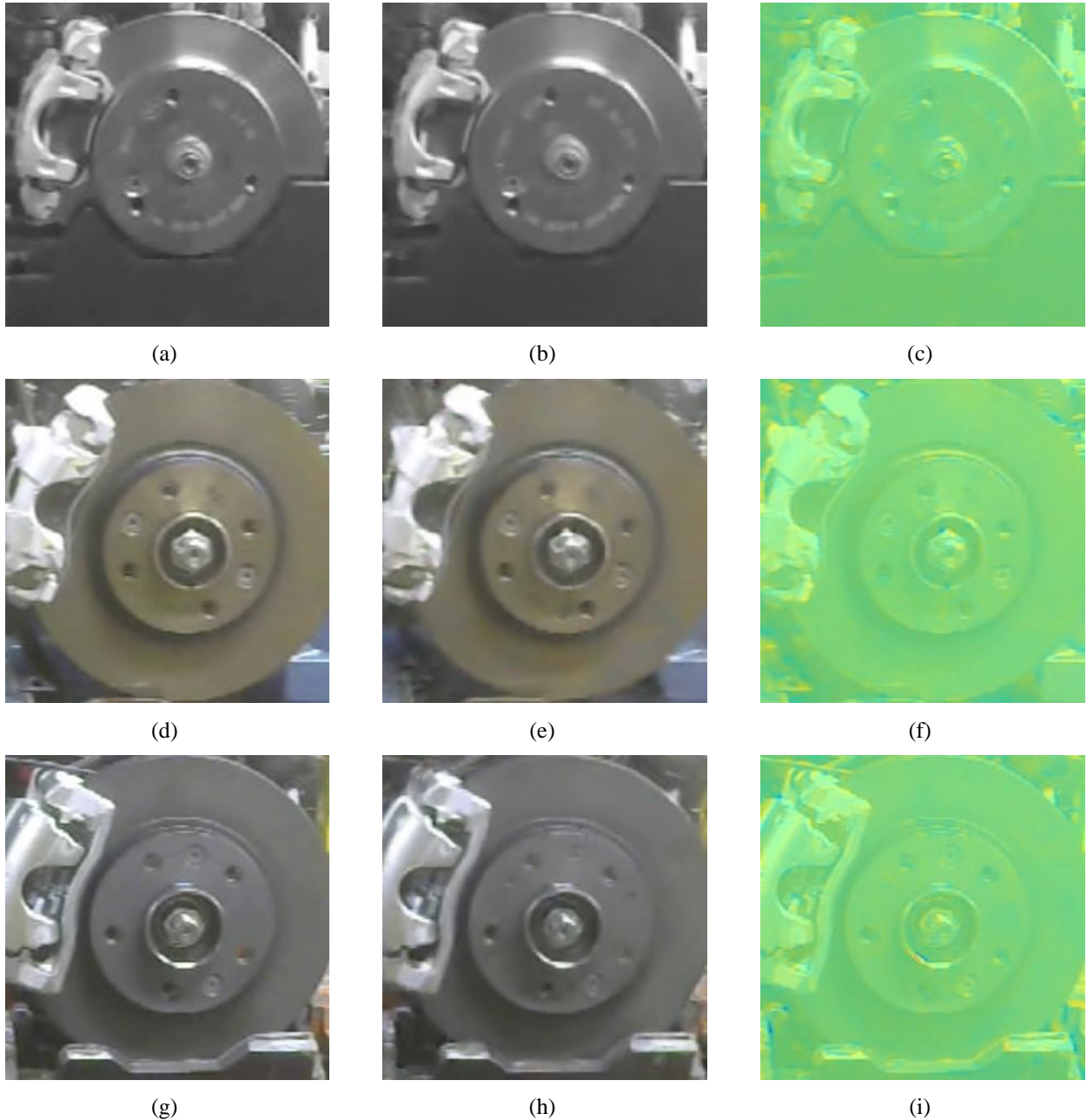

*Fig.20: Anomaly detection in normal images. (a), (d) and (g) are the real captured images. (b), (e) and (h) are generated image from respective input. (c), (f) and (i) are the residual image with colormap applied over original input.*

One contribution that the chaining of the models provides is the more detailed analysis directed to each part, since each component has a specialized model trained to detect their respective anomalies in the production process. In addition, the execution remained restrictted to the work cycle time without interfering with the environment.

However, due to the camera's positioning and resolution, it was not possible to detect scratches and bumps in the anomaly detection process. These situations can be re-evaluated by improving the used camera and resizing the architecture parameters such as the input dimensions and the latent space vector dimension.

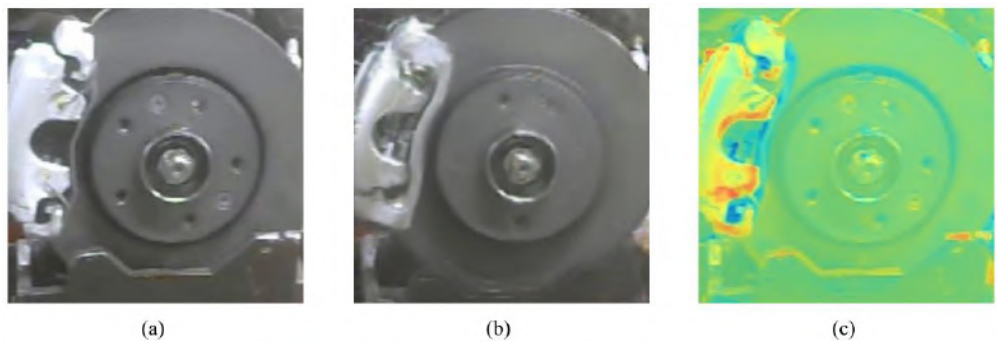

*Fig.21: Anomaly detection in brake kit with different calliper type. (a) is the real captured images. (b) is generated image from respective input. (c) is the residual image with colormap applied over original input.*

## IV. CONCLUSIONS ABOUT DEEP LEARNING APPLICATIONS IN A MANUFACTURING LINE

Experimental results indicate that inspection produced with deep learning methods are capable of working in more generalist and less constrained environments.

Furthermore, we conclude that object detection is a great tool for assisting with process quality, evaluating the result of the operational sequence of an assembly line. Semantic segmentations, on the other hand, facilitate the visualization of specific regions of parts, such as creating a mask to evaluate areas. At last, anomaly detection provides a robust unsupervised learning tool that assists in product quality assessment by verifying parts at completion of production.

Moreover, the chaining of the different methodologies makes it possible to evaluate both processes and product quality without interfering with the production line cycle. Besides, the main feature common to all addressed methodologies is the end-to-end tool aspect. That is, non-experts with low understanding of machine learning or feature extraction are able to train and apply classification models.

A disadvantage of these deep learning systems is the computational power that is demanded in real time executions, which requires an investment in robust cameras, GPUs and CPUs. An alternative to the costly solution is cloud processing or local servers, which in turn may exhibit higher response latency and require complex installation.

Future work involve exploring mechanisms for integrating production line systems with distributed deep learning [35] and other tasks for deep learning models like content-based image retrieval with image captioning [36] and biometrics analysis [37] for operator safety purposes. Besides, other proofs of concept can be explored by applying reinforcement learning [10] to mimic the dynamic behavior of operators.